\documentclass{article}
\usepackage[style=apa, backend=biber]{biblatex} 
\DeclareLanguageMapping{british}{british-apa} 
\DeclareFieldFormat[article]{volume}{\apanum{#1}} 

\usepackage{threeparttable}
\usepackage{multirow}

\usepackage{commath}
\usepackage{pdflscape}

\usepackage[letterpaper,top=2cm,bottom=2cm,left=3cm,right=3cm,marginparwidth=1.75cm]{geometry}
\usepackage{threeparttable}

\usepackage{amsmath}
\usepackage{graphicx}
\usepackage[colorlinks=true, allcolors=blue]{hyperref}
\usepackage{import}
\usepackage{subfig}
\usepackage{wasysym}
\usepackage{fontawesome5}
\usepackage{rotating}
\usepackage{adjustbox}
\usepackage{float} 
\usepackage{caption}  
\usepackage{capt-of} 
\usepackage{authblk}
\usepackage{calc}
\usepackage{array}
\usepackage{adjustbox}   
\newlength{\remaining}
\newcommand{\showremaining}{
  \setlength{\remaining}{\pagegoal-\pagetotal}
  Remaining vertical space on page: \the\remaining\ (about \the\dimexpr\remaining\relax)
}
\usepackage{authblk}
\usepackage{orcidlink}

\title{Ankh3: Multi-Task Pretraining with Sequence Denoising and Completion Enhances Protein Representations}

\author{Hazem Alsamkary\,\orcidlink{0009-0006-6966-9687}}
\author{Mohamed Elshaffei\,\orcidlink{0009-0008-4484-1668}}
\author{Mohamed Elkerdawy}
\author{Ahmed Elnaggar\thanks{Correspondence: \texttt{publications@proteinea.com}}}

\affil{Proteinea Inc.}
\date{}

\begin{document}
\setlength{\parindent}{0pt}
\maketitle
\begin{abstract}
\noindent Protein language models (PLMs) have emerged as powerful tools to detect complex patterns of protein sequences. However, the capability of PLMs to fully capture information on protein sequences might be limited by focusing on single pre-training tasks. Although adding data modalities or supervised objectives can improve the performance of PLMs, pre-training often remains focused on denoising corrupted sequences. To push the boundaries of PLMs, our research investigated a multi-task pre-training strategy. We developed Ankh3, a model jointly optimized on two objectives: masked language modeling with multiple masking probabilities and protein sequence completion relying only on protein sequences as input. This multi-task pre-training demonstrated that PLMs can learn richer and more generalizable representations solely from protein sequences. The results demonstrated improved performance in downstream tasks, such as secondary structure prediction, fluorescence, GB1 fitness, and contact prediction. The integration of multiple tasks gave the model a more comprehensive understanding of protein properties, leading to more robust and accurate predictions.
\end{abstract}

\section{Introduction}
Protein language models (PLMs) have led to a paradigm shift in the field of synthetic biology, enabling effective modeling of diverse tasks and the generation of novel proteins. Current language models typically undergo training on a single task, masked language modeling (MLM) utilizing bi-directional encoder models \parencite{rives2019biological}, encoder-decoder models \parencite{elnaggar2023ankh,elnaggar2021prottrans}, or decoder-only models \parencite{madani2020progen}. In the present research, we introduce Ankh3, a PLM pre-trained on only two tasks out of three from the UL2 objective \parencite{tay2022ul2}. The UL2 objective contains three tasks: R-denoiser [NLU] that focuses on high corruption of short spans, S-denoiser [S2S] that is a sequential denoiser, and  X-denoiser [NLG] that focuses on extreme denoising of a mix of short and long spans with high and low denoising. X-denoiser was not used due to previous experiences of long span masking in former Ankh models that resulted in poor performance. Hence, this research only focuses on short-span sequence variable denoising/demasking [NLU] and sequence completion/sequential denoising [S2S]. Sequence denoising aligns with the regular T5 denoising objective in which sentinel tokens are incorporated and sequence completion is modeled as completing the remaining half of a given protein sequence presented to the model. Two sizes of Ankh3 are provided (Table \ref{table:tab_1}), Ankh3-Large and Ankh3-XL having the following configurations: 
\clearpage
\begin{center}
\captionof{table}{Overview of the main model parameters and design choices}
\vspace{1mm}
\begin{tabular}{c c c}
\label{table:tab_1}

 Parameters & Large & XL \\ [0.5ex]
 \hline
 Embedding dim & 1536 & 2560 \\
 Layers & 72 & 72 \\
 Encoder layers & 48 & 48 \\
 Decoder layers & 24 & 24 \\
 Tie word embeddings & no & no \\
 \hline
 Feedforward dim & 3840 & 6720 \\
 Non-linearity & SwiGLU & SwiGLU \\
 \hline
 Num heads & 16 & 32 \\
 KV dim & 64 & 64 \\
 Relative attention max distance & 128 & 128 \\
 Relative attention num buckets & 64 & 64 \\
  \hline
 Num sentinel tokens & 225 & 225 \\
 Vocab size & 256 & 256 \\
 \hline
\end{tabular}
\end{center}

\section{Pre-training Configuration}
\subsection{Modeling}
\label{section_2p1}
In contrast to using the conventional masked language modeling method, as implemented in Ankh \parencite{elnaggar2023ankh}, ESM2 \parencite{rives2019biological} and ESM3 \parencite{hayes2024simulating} models or autoregressive models as implemented in ProGen2 \parencite[]{nijkamp2022progen2exploringboundariesprotein}, we added another pre-training task where the model is tasked to complete the remaining 50\% of an input sequence. The first 50\% of the sequence was given to the model as input to the encoder, and the remaining 50\% of the sequence was generated by the decoder conditioned on the output representation of the first half. This approach aims to enhance the understanding of the models for protein sequences in terms of generation and performance on downstream tasks. Although variable completion percentages may enhance the performance of the model, it was not experimented in this work due to the limited computation power. However, in inference, the model can generate the whole protein sequence with a variable sequence length input. Finally, because one more task was added during the pre-training, the increase in the capacity of Ankh3-XL model did not lead to performance degradation as in ProtT5 \parencite[]{elnaggar2021prottrans} when scaled from ProtT5-XL to ProtT5-XXL while all hyperparameters were fixed, including the dataset.

\subsection{Architecture}
As in preceding Ankh models, Ankh3 is based on T5 architecture \parencite{2020t5} with the model configuration mentioned in table \ref{table:tab_1}, an encoder-decoder transformer model with T5 relative positional bias. Table \ref{table:tab_2} presents the total number of parameters for each model.

\begin{center}
\captionof{table}{Number of parameters for each Ankh3 model}
\vspace{1mm}
\begin{tabular}{c c c c} 
\label{table:tab_2}
 Model & Encoder parameters & Decoder Parameters & Total Parameters\\ [0.5ex]
 \hline
 Ankh3-Large & 1,151,879,680 & 727,169,536 & 1,879,049,216 \\
 Ankh3-XL & 3,484,799,488 & 2,246,107,648 & 5,730,907,136 \\

 \hline
\end{tabular}
\end{center}

\subsection{Pre-training Data}

Inspired by its predecessors Ankh \parencite{elnaggar2023ankh} and ProtTrans \parencite{elnaggar2021prottrans}, the protein language model Ankh3 underwent pre-training with a substantial dataset comprising 59,130,945 distinct sequences sourced from UniRef50. The UniProt Reference Clusters (UniRef) are a collection of databases that offer comprehensively clustered sets of protein sequences from the UniProt Knowledgebase (UniProtKB), which includes isoforms, along with chosen records from UniParc as detailed in \parencite[]{suzek2015uniref}. \\
This clustering methodology employs varying sequence similarity thresholds to ensure that the resulting clusters are non-redundant and maintain homogeneity among the sequences within each cluster. Specifically, in the UniRef100 database, every entry or cluster consists of identical sequences, including sub-fragments, irrespective of the organism from which they originate. Subsequently, the UniRef50 database is created by further clustering the representative sequences from UniRef90 (which is built by clustering UniRef100 sequences at a 90\% identity threshold) using a 50\% sequence identity cutoff \parencite[]{suzek2015uniref}. This hierarchical approach ensures a reduced redundancy dataset ideal for training large-scale protein models. \parencite{suzek2015uniref}.

\subsection{Compute Infrastructure}
Both Ankh3 models were trained utilizing TPUv4 with 64 chips. The T5x library \parencite{roberts2022t5x}, developed in JAX \parencite[]{jax2018github}, served as the implementation platform. Ankh3-Large was trained without replication or sharding, while Ankh3-XL employed both 2-way data sharding and 2-way model sharding. Optimizer states were not sharded in either model.

\subsection{Training Setup}

While conventional masked language modeling traditionally utilizes a single masking probability, the training here employed three distinct masking probabilities: 15\%, 20\%, and 30\%. These probabilities governed the proportion of masked tokens and were sampled uniformly to ensure that each masking level had an equal chance of selection in every training step. This uniform distribution facilitated the exposure of the model to a varied range of masking conditions. In the previous Ankh model \parencite[]{elnaggar2023ankh}, different denoising probabilities were explored before reaching the final masking probability, and it was concluded that some tasks performed better when Ankh was trained on higher denoising probabilities, while other tasks performed better with lower denoising probabilities. Therefore, MLM was utilized here with multiple masking probabilities during pre-training so that a good performance is achieved in most of the tasks. For the multi-task training setup, each sequence from the dataset was randomly assigned to either the MLM or sequence completion task each time it was selected for a training batch. This random assignment strategy ensured that when a sequence was sampled multiple times over the course of training, the model could be exposed to that same sequence under different task objectives—for instance, processing it for MLM in one training step and for sequence completion in another. This allowed the model to learn diverse aspects from the identical input data by experiencing it through varied task contexts. In sequence completion, the sequence was divided into two segments: the first half was given to the encoder as input, and the second segment was predicted by the decoder, conditioned on the representation of the first segment that was output by the encoder model. The training hyperparameters for each model are shown in Table \ref{table:tab_3}.

\begin{center}
\captionof{table}{Training hyperparameters of the Ankh3 models}
\vspace{1mm}
\begin{tabular}{c c c}
\label{table:tab_3}

 Hyperparameters & Ankh3 Large & Ankh3-XL \\ [0.5ex]
 \hline
 Sequence length & 512 & 512 \\
 Masking probability & 15\% 20\% 30\% & 15\% 20\% 30\% \\
 Warmup scheduler & rsqrt\_decay & rsqrt\_decay \\
 Warmup steps & 10K & 10K \\
 Learning rate & 1e-2 & 1e-2 \\
 Optimizer & Adafactor & Adafactor \\
 Num steps & $\approx$ 4M & $\approx$ 5M \\
 Batch size & 1024 & 1024 \\
 Weight decay & 0.0 & 0.0 \\
\end{tabular}
\end{center}

\section{Downstream Tasks Evaluation}
The performance of the two Ankh3 models was assessed based on four benchmarking tasks: secondary structure prediction, fluorescence prediction, GB1, and contact prediction. The benchmark settings were freezing the backbone model, extracting the representation of each sequence, pooling the representation using average pooling in sequence prediction tasks (e.g., fluorescence prediction and GB1), and finally, these representations are passed to a ConvBERT model \parencite{jiang2020convbert}. ConvBERT hyperparameters were fixed across all tasks. The hyperparameters of the ConvBERT model are summarized in Table \ref{table:tab_4}.

\begin{center}
\captionof{table}{ConvBERT hyperparameters during benchmarking}
\vspace{1mm}
\begin{tabular}{c c}
\label{table:tab_4}
Hyperparameters & ConvBERT \\
\hline
Num layers & 1 \\
Feedforward dim & Embedding dim / 2 \\
Num heads & 4 \\
Dropout & 0.1 \\
Kernel size & 7 \\
Pooling (if needed) & Global average pooling \\
Learning rate & 1e-2 \\
Warmup steps & 1000 \\
Num epochs & 20 \\
Gradient accumulation steps & 16 \\
Batch Size & 1 \\
Weight Decay & 0.0 \\
\end{tabular}
\end{center}

\begin{itemize}
    \item \textbf{Secondary structure prediction:}
Secondary structures are the conformational arrangements of the polypeptide backbone in either $\alpha$-helix, $\beta$-strand, or coils \parencite{ma2018protein}. This task gives insights into the quality of the protein sequence representation of the model being tested. It has two levels of granularity: the first level involves predicting one of three states of the secondary structure (SSP-3) for each amino acid, and the second level involves predicting one of eight states (SSP-8) of the secondary structure for each amino acid, which is more challenging. Accuracy was used as the primary metric to measure the proportion of correctly predicted states. To test the model's performance, CASP12 \parencite[]{https://doi.org/10.1002/prot.25423} and CASP14 \parencite[]{https://doi.org/10.1002/prot.26237} were used, and DSSP was used to compute the labels for each sequence.

    \item \textbf{Fluorescence prediction:}
Fluorescence is a phenomenon where certain proteins, like Green Fluorescent Protein (GFP), emit light after absorbing it. The light intensity can be highly sensitive to the amino acid sequence due to the change in structure and the efficiency of the light-emitting machinery \parencite{sample2009structure}. Each PLM was trained on a fluorescence dataset \parencite[]{rao2019evaluatingproteintransferlearning}, where the model is tasked to map a protein sequence to its corresponding fluorescence value. The fluorescence value is a single continuous value; hence, all sequence information was aggregated into a single vector using global average pooling. Spearman correlation was used as the primary metric to measure how well the model is able to order the sequences based on their predicted fluorescence.

    \item \textbf{GB1 fitness:}
GB1 is the binding part of Protein G that binds to immunoglobulins, and it is important for purifying antibodies \parencite{sommer2012fast}. The study of mutations in GB1 is considered the benchmark for understanding how mutations interact in non-additive ways, a phenomenon known as epistasis \parencite[]{10.7554/eLife.16965}. This task uses regression to assess how well variants of the GB1 protein bind after mutations were introduced at four specific locations. The GB1 dataset for this task was sourced using the FLIP benchmark \parencite[]{Dallago2021.11.09.467890}. The model aims to predict a continuous fitness score for a given protein sequence. To achieve this, information from the protein sequence as understood by the model was condensed into a single vector by global average pooling. The performance of the model was assessed using Spearman correlation to indicate how accurately the model can rank the protein sequences according to their predicted GB1 binding fitness.

    \item \textbf{Contact prediction:}
 Contact prediction assesses the ability of the model to infer the spatial proximity between pairs of amino acids from their sequence. Contacts were defined based on the Euclidean distance between the C$_{\alpha}$ atoms of residue pairs. A contact is assumed if this C$_{\alpha}$-C$_{\alpha}$ distance is less than 8.0\,\AA ngstr\"{o}ms, resulting in an $N \times N$ binary contact map for each protein of length $N$. For this task, ProteinNet \parencite[]{ProteinNet19} and CASP14 \parencite[]{https://doi.org/10.1002/prot.26237} were used. For evaluation, the predicted contacts between residue pairs $(i,j)$ were specifically considered with a sequence separation of $|i-j| \ge 6$. Performance for these contacts was assessed using precision. The primary metrics were Precision@L (P@L) and Precision@L/5 (P@L/5), where $L$ is the sequence length. P@L measures the precision among the $L$ highest-scoring predicted contacts satisfying this separation criterion, while P@L/5 evaluates this precision for the top $L/5$ such contacts.
\end{itemize}

\subsection{Evaluation Details}
The performance of the PLMs was tested in two different settings: the first setting is concatenating the [NLU] token at the beginning of each sequence, and the second setting is concatenating the [S2S] token at the beginning of each sequence. Table \ref{table:tab_5} illustrates the average performance of each model with both [NLU] and [S2S] tokens. Three different seeds (7, 0, and 42) were run for each task, and the average performance was reported. The results of the previous Ankh and ESM2 models were concluded from the original paper \parencite[]{elnaggar2023ankh}.

\begin{table}[h!]
\renewcommand{\arraystretch}{1.5}
\centering
\caption{Average performance of models is reported in percentage (\%). For Ankh3 models, [NLU] and [S2S] tokens are concatenated at the beginning of the sequence. The exceptions to percentage-based reporting are Fluorescence (FL) and GB1, for which the reported metric is Spearman correlation. Other metrics include accuracy for SSP and Precision@K for CP. SSP: Secondary structure prediction; FL: Fluorescence; CP: Contact prediction. Models under the protein sequence only input category were pre-trained using only protein sequences, while multimodal models utilized protein sequences and other modalities during pre-training (e.g. secondary structure and structure tokens).}
\resizebox{\columnwidth}{!}{%
\begin{threeparttable}
\begin{tabular}{ccc|ccccccccc|c}
\hline
\multicolumn{1}{r}{} &
  \multicolumn{1}{l}{} &
  \multicolumn{1}{l|}{} &
  \multicolumn{9}{c|}{Protein Sequence Only Input} &
  Multimodal \\ \hline
 &
   &
   &
  \multicolumn{4}{c}{Multi-task} &
  \multicolumn{1}{c|}{} &
  \multicolumn{4}{c|}{Single-task} &
  Multi-task \\ \hline
Task &
  Dataset &
   &
  \multicolumn{2}{c}{Ankh3-L} &
  \multicolumn{2}{c}{Ankh3-XL} &
  \multicolumn{1}{c|}{} &
  Ankh Base* &
  Ankh Large* &
  ESM2-650M* &
  ESM2-15B* &
  ESM3-open \\ \cline{4-13} 
 &
   &
   &
  NLU &
  S2S &
  NLU &
  S2S &
  \multicolumn{1}{c|}{} &
   &
   &
   &
   &
   \\ \hline
SSP-3 &
  CASP-12 &
   &
  78.03 ± 0.11 &
  75.49 ± 0.21 &
  \textbf{84.40 ± 0.05} &
  83.76 ± 0.10 &
  \multicolumn{1}{c|}{} &
  80.81 &
  83.59 &
  82.43 &
  83.16 &
  83.43 ± 0.02 \\
SSP-8 &
  CASP-12 &
   &
  65.29 ± 0.05 &
  62.74 ± 0.22 &
  72.53 ± 0.17 &
  72.25 ± 0.16 &
  \multicolumn{1}{c|}{} &
  68.85 &
  71.69 &
  70.50 &
  71.17 &
  \textbf{73.50 ± 0.37} \\
SSP-3 &
  CASP-14 &
   &
  79.28 ± 0.15 &
  77.96 ± 0.09 &
  82.19 ± 0.13 &
  82.30 ± 0.35 &
  \multicolumn{1}{c|}{} &
  76.67 &
  77.48 &
  76.97 &
  76.56 &
  \textbf{83.20 ± 0.16} \\
SSP-8 &
  CASP-14 &
   &
  65.50 ± 0.22 &
  65.88 ± 0.10 &
  69.85 ± 0.09 &
  69.51 ± 0.30 &
  \multicolumn{1}{c|}{} &
  62.33 &
  63.17 &
  62.10 &
  61.81 &
  \textbf{71.70 ± 0.11} \\
FL &
  TAPE &
   &
  64.56 ± 0.23 &
  64.89 ± 0.36 &
  64.23 ± 0.48 &
  \textbf{65.43 ± 0.30} &
  \multicolumn{1}{c|}{} &
  62.0 &
  62.0 &
  48.0 &
  56.0 &
  58.78 ± 0.57 \\
GB1 &
  FLIP &
   &
  90.30 ± 0.39 &
  89.44 ± 0.77 &
  89.62 ± 1.55 &
  \textbf{90.44 ± 0.89} &
  \multicolumn{1}{c|}{} &
  85.0 &
  84.0 &
  82.0 &
  57.0 &
  82.91 ± 2.53 \\
CP &
  ProteinNet (L/1) &
   &
  46.35 ± 0.49 &
  43.39 ± 0.44 &
  \textbf{60.95 ± 0.35} &
  60.76 ± 0.21 &
  \multicolumn{1}{c|}{} &
  43.21 &
  48.93 &
  29.36 &
  31.62 &
  56.46 ± 8.78 \\
CP &
  ProteinNet (L/5) &
   &
  69.42 ± 0.76 &
  66.60 ± 0.87 &
  \textbf{83.89± 0.47} &
  83.31 ± 1.21 &
  \multicolumn{1}{c|}{} &
  66.63 &
  73.49 &
  50.74 &
  52.97 &
  77.40 ± 5.80 \\
CP &
  CASP-14 (L/1) &
   &
  16.94 ± 0.56 &
  18.59 ± 0.31 &
  \textbf{29.23 ± 0.36} &
  28.68 ± 0.31 &
  \multicolumn{1}{c|}{} &
  13.50 &
  16.01 &
  13.71 &
  14.44 &
  28.65 ± 1.38 \\
CP &
  CASP-14 (L/5) &
   &
  26.65 ± 0.49 &
  29.96 ± 2.09 &
  46.17 ± 1.72 &
  47.30 ± 2.09 &
  \multicolumn{1}{c|}{} &
  28.65 &
  29.91 &
  22.25 &
  26.61 &
  \textbf{47.35 ± 4.60}
\end{tabular}%
\begin{tablenotes}
      \item[*] Sourced from the Ankh paper; standard deviation is not provided here as the original publication reported error bars for these specific entries instead.
    \end{tablenotes}
      \end{threeparttable}
}
\label{table:tab_5}
\end{table}
\section{Discussion}
While NLU and S2S tasks are different, neither of them consistently performed better than the other across all the tasks and model sizes. The preferred objective seems to be task-dependent. One interesting observation that requires deeper investigation is that Ankh3-XL performed better with S2S in sequence classification tasks, such as GB1 and fluorescence (Table \ref{table:tab_5}). However, this pattern should be tested with more tasks to confirm its consistency.\\

In the context of scaling Ankh3-Large to Ankh3-XL, Ankh3-XL performed significantly better in both NLU and S2S tasks, indicating that the capacity of Ankh3-Large was insufficient to handle both tasks. As previously discussed in Section \ref{section_2p1} and demonstrated by experiments in ProtT5 \parencite{elnaggar2021prottrans}, no improvement in performance is achieved by solely increasing the model size and using protein sequences as the only input while keeping all other factors constant (including the dataset), for example, CASP12 reached 84.4\% in Ankh3-XL compared to 79.2\% in ProtT5-XXL (4.8B encoder parameters) and 81.4\% in ProtT5-XL (1.2B encoder parameters), this indicates that the addition of multiple masking probabilities and sequence completion were the main contributors to the performance boost while using protein sequences as the only input. Ankh3 performed better in all tasks compared to protein sequence-only models like ESM2 and Ankh. When compared to ESM3, Ankh3 has competitive performance on tasks such as SSP and CP. However, in tasks that neither Ankh3 nor ESM3 encountered during pre-training, Ankh3-XL performed significantly better, which indicates that Ankh3-XL may have better generalization, especially in tasks like fluorescence and GB1 fitness prediction.
\\ \\
ESM3 was primarily trained on masked language modeling, which processes multiple discrete inputs and outputs; these inputs included secondary structure tokens with eight states, structure tokens, and other different modalities. Since ESM3 was already trained with such secondary structure inputs, it was expected to outperform Ankh3 in tasks evaluating this feature, as reported in Table \ref{table:tab_5}. It was also hypothesized that ESM3 would outperform Ankh3 in contact prediction, as its training included structure tokens that can enhance contact prediction accuracy. This hypothesis was confirmed with seeds 7 and 0, where ESM3 performed noticeably better than Ankh3. However, with seed 42, ESM3's performance was significantly lower and also exhibited a large standard deviation (Table \ref{table:tab_5}). For example, when run with seed 42 on ProteinNet (L/1), ESM3 showed a standard deviation of 8.78\%. This high variability observed with seed 42 likely contributed to ESM3's overall poorer performance metrics when compared to Ankh3, which demonstrated stability across all three seeds in all tasks.

\section{Conclusion}
In this work, we presented Ankh3, the latest version of the Ankh family. Ankh3 performance was demonstrated in the setting of multi-task pre-training. Sequence completion was explored as an additional task, and masked language modeling with multiple masking probabilities was shown to enhance the performance of the model. These results highlight the continued potential for advancing sequence-only protein language models through innovative multi-task learning, yielding more robust and versatile protein representations without immediate reliance on additional data modalities. To foster reproducibility and further research, we shared all of our work and details, including pre-training hyperparameters, along with the dataset on Huggingface. Finally, the weights of both Ankh3 models are available open source on Huggingface. Our future research directions include scaling Ankh3 to incorporate multiple modalities and further exploring sequence completion with variable completion percentages. \\

\section{Data Availability}
The dataset used for pre-training is available at \url{https://huggingface.co/datasets/agemagician/uniref50}. 
Model weights of both Ankh3-Large and Ankh3-XL models are available at \url{https://huggingface.co/ElnaggarLab/ankh3-large} and \url{https://huggingface.co/ElnaggarLab/ankh3-xl}.

\subsection*{Acknowledgment}
The authors gratefully acknowledge the significant contributions of Proteinea's team, especially Nehal Adel Abdelsalam, for the invaluable guidance and for ensuring that the quality of this research paper meets the standards. We also gratefully acknowledge Proteinea's deep learning and bioinformatics teams, who provided essential assistance with hardware, software, and numerous other project facets. We are indebted to Google for their comprehensive support, including Jonathan Caton, Shira Genauer, Astitva Chopra, and the Google Cloud, Google Innovator, JAX, and TRC Teams, for their help in configuring the project on Google Cloud and troubleshooting technical challenges. This research was also financially supported by Google through the Google Research Innovator and Google TPU Cloud Research Credits Programs. We also extend our thanks to the HuggingFace team, particularly Patrick von Platen, Julien Chaumond, and Clement Delangue, whose support was crucial for making the trained models publicly accessible. Lastly, we express our gratitude to the global research community for making the datasets utilized in this study freely available.
\printbibliography 


\end{document}